\documentclass{ieeeaccess}
\usepackage{amsmath,amssymb,amsfonts}
\usepackage{algorithmic}
\usepackage{graphicx}
\usepackage{textcomp}
\usepackage{bm}
\usepackage{booktabs}
\usepackage[caption=false, font=footnotesize, margin=.5cm]{subfig}
\usepackage{float}
\usepackage{siunitx}

\makeatletter
\AtBeginDocument{\DeclareMathVersion{bold}
\SetSymbolFont{operators}{bold}{T1}{times}{b}{n}
\SetSymbolFont{NewLetters}{bold}{T1}{times}{b}{it}
\SetMathAlphabet{\mathrm}{bold}{T1}{times}{b}{n}
\SetMathAlphabet{\mathit}{bold}{T1}{times}{b}{it}
\SetMathAlphabet{\mathbf}{bold}{T1}{times}{b}{n}
\SetMathAlphabet{\mathtt}{bold}{OT1}{pcr}{b}{n}
\SetSymbolFont{symbols}{bold}{OMS}{cmsy}{b}{n}
\renewcommand\boldmath{\@nomath\boldmath\mathversion{bold}}}
\makeatother

\def\BibTeX{{\rm B\kern-.05em{\sc i\kern-.025em b}\kern-.08em
    T\kern-.1667em\lower.7ex\hbox{E}\kern-.125emX}}

\begin{document}
\history{Date of publication xxxx 00, 0000, date of current version xxxx 00, 0000.}
\doi{}

\title{Joint Analysis of Optical and SAR Vegetation Indices for Vineyard Monitoring: Assessing Biomass Dynamics and Phenological Stages over Po Valley, Italy}

\author{\uppercase{Andrea Bergamaschi}\authorrefmark{1},
\uppercase{Abhinav Verma}\authorrefmark{2}, \IEEEmembership{Member, IEEE}, \uppercase{Avik Bhattacharya}\authorrefmark{2},
\IEEEmembership{Senior Member, IEEE}, and \uppercase{Fabio Dell'Acqua}\authorrefmark{1} \IEEEmembership{Senior Member, IEEE}}

\address[1]{Department of Electrical, Computer, Biomedical Engineering, University of Pavia, Pavia, Italy (e-mail: fabio.dellacqua@unipv.it)}
\address[2]{Centre of Studies in Resources Engineering (CSRE), Indian Institute of Technology Bombay, Mumbai, India (e-mail: avikbhat@iitb.ac.in)}
\tfootnote{This research was partly funded by the European Union - NextGenerationEU, Mission 4 Component 1.5 - ECS00000036 - CUP F17G22000190007.}


\corresp{Corresponding author: Fabio Dell'Acqua (e-mail: fabio.dellacqua@unipv.it).}

\begin{abstract}
Multi-polarized Synthetic Aperture Radar (SAR) technology has gained increasing attention in agriculture, offering unique capabilities for monitoring vegetation dynamics thanks to its all-weather, day-and-night operation and high revisit frequency. This study presents, for the first time, a comprehensive analysis combining dual-polarimetric radar vegetation index (DpRVI) with optical indices to characterize vineyard crops. Vineyards exhibit distinct non-isotropic scattering behavior due to their pronounced row orientation, making them particularly challenging and interesting targets for remote sensing. The research further investigates the relationship between DpRVI and optical vegetation indices, demonstrating the complementary nature of their information. We demonstrate that DpRVI and optical indices provide complementary information, with low correlation suggesting that they capture distinct vineyard features. Key findings reveal a parabolic trend in DpRVI over the growing season, potentially linked to biomass dynamics estimated via the Winkler Index. Unlike optical indices reflecting vegetation greenness, DpRVI appears more directly related to biomass growth, aligning with specific phenological phases. Preliminary results also highlight the potential of DpRVI for distinguishing vineyards from other crops. This research aligns with the objectives of the PNRR-NODES project, which promotes nature-based solutions (NbS) for sustainable vineyard management. The application of DpRVI for monitoring vineyards is part of integrating remote sensing techniques into the broader field of strategies for climate-related change adaptation and risk reduction, emphasizing the role of innovative SAR-based monitoring in sustainable agriculture.
\end{abstract}


\titlepgskip=-21pt

\maketitle

\section{Introduction}
\label{sec:introduction}
\PARstart{V}{iticulture} plays a crucial role in the agricultural economy of many countries, supporting local markets, tourism, and cultural heritage. In the European Union (EU), for example, as of 2020, the total area under vines was 3.2 million hectares (\SI{}{\hectare}), equivalent to \SI{2.0}{\percent} of the total managed agricultural area. Spain, France, and Italy accounted for three-quarters (\SI{74.9}{\percent}) of the total area under vines~\cite{eurostat}. To increase vineyard production, new technologies were introduced in the late 1990s, shifting from classical viticulture to “precision viticulture” (PV)~\cite{arno2009precision}. PV manages vineyards at the parcel level, improving yield prediction and harvest efficiency. In this context, remote sensing (RS) is a fundamental tool that allows data to be collected without physical contact. 

Most RS applications for vineyards were developed using high-resolution optical imagery and high-precision UAV (Unmanned Aerial Vehicle) data. However, these technologies have some limitations, such as their dependence on Sun illumination and atmospheric conditions, which can lead to the inability to consistently and efficiently acquire data. These constraints can be overcome with Synthetic Aperture Radar (SAR) technology~\cite{bakon2024synthetic}. SAR sensors can operate regardless of sunlight or atmospheric conditions, offer short revisit times, and penetrate canopies to sense soil backscatter. Unlike optical sensors, SAR is sensitive to target dielectric and geometric properties, revealing the structural characteristics of the target~\cite{bhattacharya2021radar}.

Several studies have used SAR technology to explore vineyard structure and morphology~\cite{schiavon2007sensitivity,loussert2016analysis,beeri2020kc,davitt2022complementary}. However, they often show significant limitations, such as small datasets, challenges in processing SAR signals, and an inability to fully address the inherent complexity of vineyard analysis. Despite these drawbacks, this type of microwave sensing could provide additional clues on the status of monitored vines, especially considering recently proposed radar-based vegetation indices.

In this study, we utilize the dual-pol Sentinel-1 SAR data to derive the Dual-pol Radar Vegetation Index (DpRVI) and analyze it jointly with multispectral-based indices. DpRVI~\cite{mandal2020dual} is based on the eigenvalue spectrum of the wave covariance matrix ($\lambda_1$, $\lambda_2$). Alternatively, it can also be expressed using the degree of polarization $m$ and the degree of dominance $\beta$ derived from the eigenvalues that may provide additional clues on vineyards. 

The three main objectives of this research are: 
\begin{itemize}
    \item To explore the correlation between the DpRVI and optical vegetation indices for vineyards.
    \item Additionally, to assess whether dual-polarimetric SAR data can complement optical datasets by filling observational gaps in vineyard monitoring.
    \item To evaluate the capability of DpRVI temporal profiles to identify and track phenological stages in vineyard ecosystems.
\end{itemize}
Although the size of the sample and its limited geographical extent constrain the scope of results, this investigation can be useful as a precursor towards joint optical and radar monitoring of vineyards at a larger scale.

\section{Raw data and tools}

\subsection{Study Area}

The vineyard blocks used in this study are located near Santa Maria della Versa (Italy), on the Oltrepò Pavese hills, in the middle valley of the Versa stream. The Oltrepò Pavese vineyard district is mainly located in hilly areas of the Pavia Province, a top grape producer in the Lombardy Region with 13,000 hectares out of a regional total of 24,600 hectares~\cite{sadeghi2024game}. According to Koppen’s classification, the climatic regime of the Oltrepò Pavese area is temperate/mesothermal~\cite{bordoni2019effects}, which makes it ideal for grape cultivation and harvest. 

The phenological cycle starts between late March and early April with sprouting. This phase is followed by leaf development through the entire month of April. From early May until June, flowering takes place. Fruit development occurs from mid-June into July, followed by veraison. The ripening of berries begins in mid-July and continues until September when the vineyard yield is ready to be harvested. Harvesting usually occurs between mid-September and early October. However, over the past few decades, the Oltrepò Pavese region has experienced a significant increase in average temperatures, with mean annual temperatures rising by approximately \SI{1.6}{\degreeCelsius} between 1961–1990 and 1991–2017. This warming trend has led to earlier phenological events, such as budburst and veraison~\cite{vercesi2024autochthonous}. 

\begin{figure}[h]
    \centering
    \includegraphics[width=0.6\linewidth]{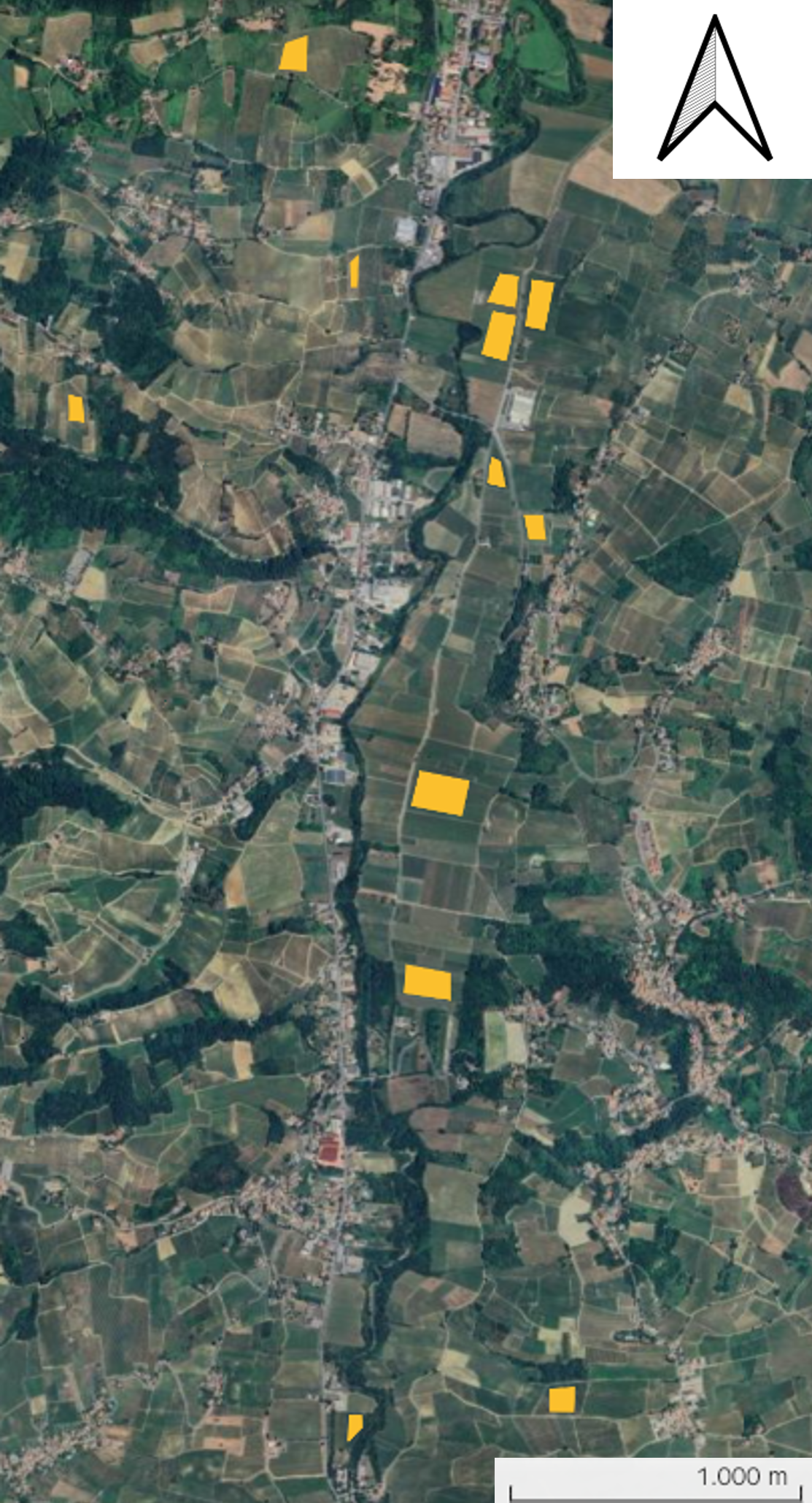}
    \caption{Vineyards analysed in this study (centre coordinates of the study area: 45°02'40"N 9°17'59"E). Twelve fields, highlighted in yellow, are analysed. Photo courtesy of Google Earth.}
    \label{fig:Parcels}
\end{figure}

In this context, a set of sample vineyards was identified and precisely outlined in GIS polygons, cross-checked by ground surveying and visual interpretation of high spatial resolution optical satellite images.
The proposed methodology was tested on $12$ vineyards arranged into two groups based on row orientation: $6$ vineyards with a predominantly East-West (EW) row orientation, referred to as “EW1” through “EW6”, and $6$ vineyards with a predominantly North-South (NS) row orientation, referred to as “NS1” through “NS6”. Vineyards with other orientations were discarded to avoid geometric bias on radar acquisition results. The Region of Interest (ROI) polygon for each vineyard was defined with a slight inward buffer to mitigate edge effects. The concerned parcels are clearly visible in the Google Earth picture shown in Figure \ref{fig:Parcels}. 

\subsection{Data sets}

A total of $12$ dual-pol Sentinel-1 (S-1) images were selected over the study area: six from an ascending satellite orbit and six from a descending satellite orbit. The image dates were chosen to highlight the different vineyard phenological phases, from early March (bud development) to late August (grape maturation). Although Sentinel-1 can acquire images every $12$ days, we selected monthly images. This approach was chosen to capture only meaningful growth variations, reduce redundancy, and optimize computational resources.

S-1 images were downloaded individually from the Copernicus Browser (https://browser.dataspace.copernicus.eu/) in the Terrain Observation with Progressive Scans SAR (TOPSAR) mode and delivered to the user as Level-1 SLC in Interferometric Wide swath (IW) acquisition mode. IW provides data with a large swath width (\SI{250}{\kilo\meter}) and with $5\times20$ m spatial resolution.

On the other hand, $6$ Sentinel-2 (S-2) datasets were acquired. These were chosen based on two criteria: they had to be cloud-free over the ROI and as near as possible in time to the corresponding Sentinel-1 acquisition. Furthermore, the Sentinel-2 datasets were selected from the same tile (specifically, tile 32TNQ) to ensure consistency regarding geographical referencing. The characteristics of the chosen images are listed in tables~\ref{tab:Sentinel-1} and~\ref{tab:Sentinel-2}.

\begin{table}[!h]
    \centering
    \caption{Characteristics of the Sentinel-1 images.}
    \resizebox{1\columnwidth}{!}{%
    \begin{tabular}{cccc}
        \toprule Acquisition date & Beam mode & Acquisition time & Orbit \\ \midrule
        $2023-03-28$ & IW & $\sim 05:25$ & Descending \\
        $2023-03-29$ & IW & $\sim 17:15$ & Ascending \\
        $2023-04-21$ & IW & $\sim 05:25$ & Descending \\
        $2023-04-22$ & IW & $\sim 17:15$ & Ascending \\
        $2023-05-27$ & IW & $\sim 05:25$ & Descending \\
        $2023-05-28$ & IW & $\sim 17:15$ & Ascending \\
        $2023-06-20$ & IW & $\sim 05:25$ & Descending \\
        $2023-06-21$ & IW & $\sim 17:15$ & Ascending \\
        $2023-07-20$ & IW & $\sim 05:25$ & Descending \\
        $2023-07-21$ & IW & $\sim 17:15$ & Ascending \\
        $2023-08-19$ & IW & $\sim 05:25$ & Descending \\
        $2023-08-20$ & IW & $\sim 17:15$ & Ascending \\
        \bottomrule
    \end{tabular}}
    \label{tab:Sentinel-1}
\end{table}

\begin{table}[!h]
    \centering
    \caption{Characteristics of the Sentinel-2 images.}
    \resizebox{0.75\columnwidth}{!}{%
    \begin{tabular}{ccc}
        \toprule Acquisition date & Platform & Tile ID \\
        \midrule 
        $2023-03-27$ & S2A & 32TNQ \\
        $2023-04-26$ & S2A & 32TNQ \\
        $2023-05-26$ & S2A & 32TNQ \\
        $2023-06-25$ & S2A & 32TNQ \\
        $2023-07-25$ & S2A & 32TNQ \\
        $2023-08-24$ & S2A & 32TNQ \\
        \bottomrule
    \end{tabular}}
    \label{tab:Sentinel-2}
\end{table}

Note that the time of each acquisition depends on the selected orbit direction, and thus, their acquisition times are different. In particular, ascending images were acquired in the afternoon (17:15), whereas descending images were acquired in the morning (05:35). The acquisition time may affect the analysis since vegetative activity changes throughout the day.

\section{Methods and materials}
This section briefly describes the SAR- and optical-derived indices used in this study. 

\subsection{SAR data preprocessing}

The dual-pol Sentinel-1 SLC SAR data used in this study were processed in ESA's SNAP 11.0 software following the standard processing steps~\cite{mandal2019sentinel}.

The following processing steps were applied to each image: 1) \texttt{S-1 TOPS Split} -- select a sub-swath of the three. 2) \texttt{Apply Orbit File} -- update satellite orbit file information. 3) \texttt{Calibrate} -- relate pixel values to the actual radar backscatter. 4) \texttt{S-1 TOPS Deburst} -- merge all bursts in the selected sub-swath. 5) \texttt{Polarimetric Matrix Generation} -- generate wave covariance $\mathbf{C}_{2}$ matrix. 6) \texttt{Multilooking} -- reduce inherent speckle and obtain square pixel 7) \texttt{Polarimetric Speckle
Filter (optional)} -- reduce speckle further, 8) \texttt{Range Doppler Terrain Correction} -- geocode the image.

Once the processed wave covariance $\mathbf{C}_{2}$ matrix was obtained, we computed the DpRVI for each date, as discussed in the following section. The multi-date DpRVI images were then stacked to ensure sub-pixel alignment.

\subsection{Dual-pol Radar Vegetation Index (D\lowercase{p}RVI)}

Radar vegetation indices serve as excellent proxies for vegetation characteristics. It offers straightforward and physically interpretable insights into vegetation structure and condition. Mandal et al.~\cite{mandal2020dual} proposed the expression of DpRVI for SLC SAR data that utilizes the scattering information from targets in terms of the eigenvalue spectrum of the $\mathbf{C}_{2}$ matrix. It is calculated as:
\begin{align}
    \text{DpRVI} &= 1 - m\ \beta, \\
    &= \dfrac{q \left(q + 3 \right)}{\left(q + 1 \right)^{2}} 
    \label{dprvi}
\end{align}
where $m = (\lambda_{1} - \lambda_{2})/(\lambda_{1} + \lambda_{2})$ is the degree of polarization, and $\beta = \lambda_{1}/(\lambda_{1} + \lambda_{2})$ is the degree of dominance~\cite{mandal2020dual}. Alternatively, it can be expressed in terms of $q$, where $q = \lambda_{2}/\lambda_{1}$. Here, $\lambda_{1}$ and $\lambda_{2}$ are the eigenvalues of the $2\times 2$ $\mathbf{C}_{2}$ matrix. Note that the DpRVI can also be computed from Ground Range Detected (GRD) dual-pol SAR data that only have the intensity information of the two polarization channels (i.e., diagonal elements of the $\mathbf{C}_{2}$ matrix). Bhogapurapu et al.~\cite{bhogapurapu2021dual} proposed the expression of DpRVI for the GRD data with $q = \sigma^{0}_{VH}/\sigma^{0}_{VV}$ in equation~\ref{dprvi}.

DpRVI ranges between $0$ and $1$, with low values close to $0$ indicating pure targets (e.g., smooth bare fields) and high values close to $1$ indicating random targets (e.g., crops at advanced growth stages). Thus, the temporal variations in the values of DpRVI can be related to the crop's phenological phases and morphological evolution. In this study, we computed DpRVI directly in the SNAP 11.0 software~\cite{mandal2019sentinel,bhogapurapu2021dual}. 

\subsection{Optical Data Preprocessing}

The optical vegetation indices used in this study are derived from Sentinel-2 Level-2A (L2A) products that provide orthorectified Bottom of Atmosphere (BOA) surface reflectance. This choice was made because L2A data are already geometrically and radiometrically corrected, making the preprocessing effort minimal. 

The processing of Sentinel-2 optical datasets included three steps: (i) subset image to area of interest (aoi), (ii) upsampling the data, and (iii) computation of vegetation indices. An image subset was performed to reduce the total size of the image and improve computational efficiency. 

Upsampling was carried out to increase the spatial resolution of the image up to \SI{10}{\meter}, as not all Sentinel-2 bands achieve this resolution. This way, the data were more homogeneous and easy to compare, which is fundamental for multi-temporal analysis. Finally, we computed two optical vegetation indices: NDVI~\cite{huang2021commentary} and SVHI~\cite{kumar2023sentinel, kumar_cropres2024} and a plant descriptor, Leaf Area Index (LAI). 

\subsection{Optical Vegetation indices}

The Normalized Difference Vegetation Index (NDVI) is a widely used parameter for assessing vegetation health and density~\cite{huang2021commentary}. Using the B4 (RED) and B8 (NIR) spectral bands of Sentinel-2 data, it is computed as:

\begin{equation}
    \mathrm{NDVI}=\frac{B8-B4}{B8+B4}  
\end{equation}

Its values range between $-1$ and $+1$, with higher values depicting dense and healthy vegetation.

NDVI is sensitive to vineyard biomass and density. In particular, NDVI was used to identify crop rows~\cite{ronchetti2020crop}, estimate biophysical and geometrical parameters \cite{caruso2017estimating}, and monitor water stress~\cite{zuniga2017high}.

NDVI is primarily sensitive to leaf chlorophyll and thus serves as an indicator of vegetation greenness. This study also includes the Sentinel-2 Vegetation Health Index (SVHI) proposed by Kumar et al.~\cite{kumar2023sentinel} sensitive to chlorophyll, leaf water, and protein content of plant leaves. The SVHI helps in detecting vegetation stress. In this present study, SVHI helped assess the seasonal rigor of the Vineyard crop.

Using the B4 (RED), B5 (RED-EDGE), B8 (NIR), B11 and B12 (SWIR) spectral bands of Sentinel-2 data SVHI is computed as:
\begin{equation}
    \mathrm{SVHI}=\frac{4 B 8-(B 4+B 5+B 11+B 12)}{4 B 8+(B 4+B 5+B 11+B 12)} 
\end{equation}
 
SVHI ranges between $-1$ and $1$, with higher values indicating healthy vegetation. The main advantage of using the SVHI index is that it is related to both chlorophyll and water content of the plant, whereas NDVI is only sensitive to chlorophyll content~\cite{dennison2005use}. This makes SVHI a more robust indicator for detecting vegetation stress and monitoring overall plant health.

Finally, a plant descriptor, LAI, was retrieved using the \texttt{Biophysical Processor} tool available in the SNAP software~\cite{weiss2020s2toolbox}. This tool is designed to extract different biophysical parameters related to crop health and structure. In particular, this processor was used to retrieve the LAI using the Sentinel-2 images. It is based on neural networks trained using radiative transfer models (RTM). It requires specific Sentinel-2 bands as input (i.e., B3, B4, B5, B6, B7, B8A, B11, B12) along with corresponding viewing and illumination angles~\cite{weiss2020s2toolbox}. The algorithm can be applied to any vegetation type since its main objective is to be adaptable for various applications while maintaining reasonable performance.

LAI is a fundamental plant's biophysical parameter that quantifies the (one-sided) leaf area per unit of terrain surface area. LAI can be used to study different vegetative phenomena such as photosynthesis, evaporation, transpiration, and carbon flux~\cite{zheng2009retrieving}. For this reason, LAI can be an essential parameter when analyzing vineyard biomass.

\subsection{Experimental analysis}

The radar-based vegetation index DpRVI is first compared to other optical-based vegetation indices and plant descriptors, viz., NDVI, SVHI, and LAI, through the development of scatter plots, temporal trends, and correlation analyses. Then, DpRVI is evaluated to assess its capability to accurately describe the evolution of the vineyard biomass trend. 

The lack of ground truth measurements of vineyard biomass or any other related biophysical parameter called for a different solution to link biomass with DpRVI evolution. 

Castelan-Estrada et al.~\cite{castelan2002allometric} developed allometric relationships that link vine structure features (such as stem diameter and length) to the biomass of different vine components. These allometric equations were used to estimate vineyard biomass during the growing season. In order to improve the accuracy of these equations and take into account the effect of temperature on crop seasonal variation, a cumulative degree days (CDD) system (also known as the Winkler Index) was introduced. 

Growing Degree days (GDD) are measurements of the accumulated heat over time and are directly linked to temperature~\cite{costa2019grapevine}. They are calculated based on the difference between daily mean temperatures and a specific baseline temperature (\SI{10}{\degreeCelsius} for vineyards), which represents the minimum temperature required for significant vegetative activity. Cumulative degree days represent the cumulative sum of degree days over the year.

\begin{figure*}[h!]
\centering
\subfloat[DpRVI EW, Asc.]{%
	\includegraphics[width=0.60\columnwidth]{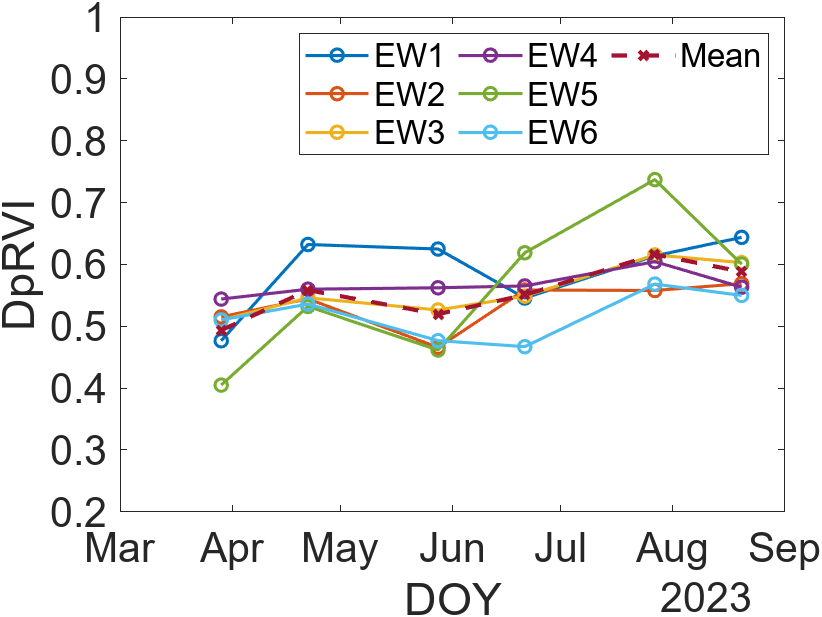}\label{PlotEW_DpRVI_asc}}
\hspace{2em}
\subfloat[DpRVI EW, Des.]{%
	\includegraphics[width=0.60\columnwidth]{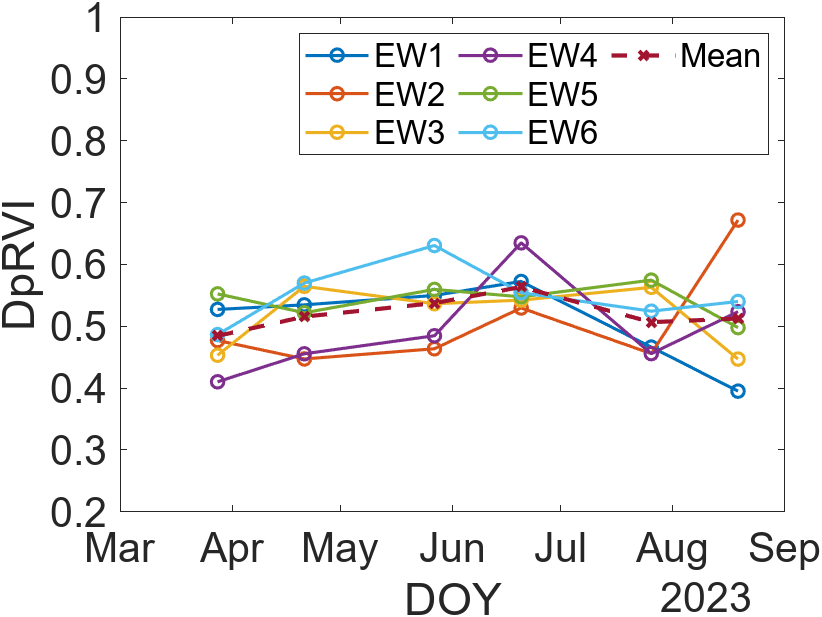}\label{PlotEW_DpRVI_des}}
\hspace{2em}
\subfloat[NDVI EW]{%
	\includegraphics[width=0.60\columnwidth]{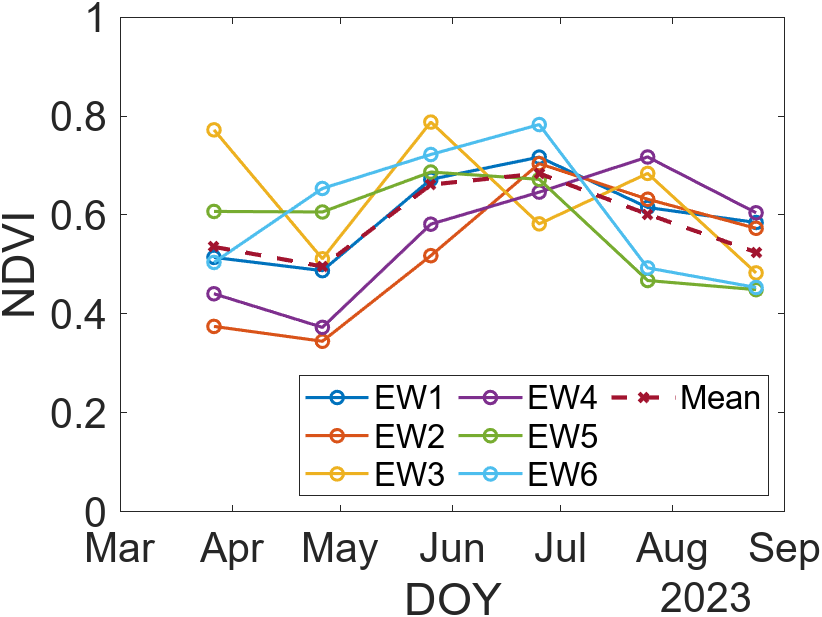}\label{PlotEW_NDVI}}
\\
\subfloat[SVHI EW]{%
	\includegraphics[width=0.60\columnwidth]{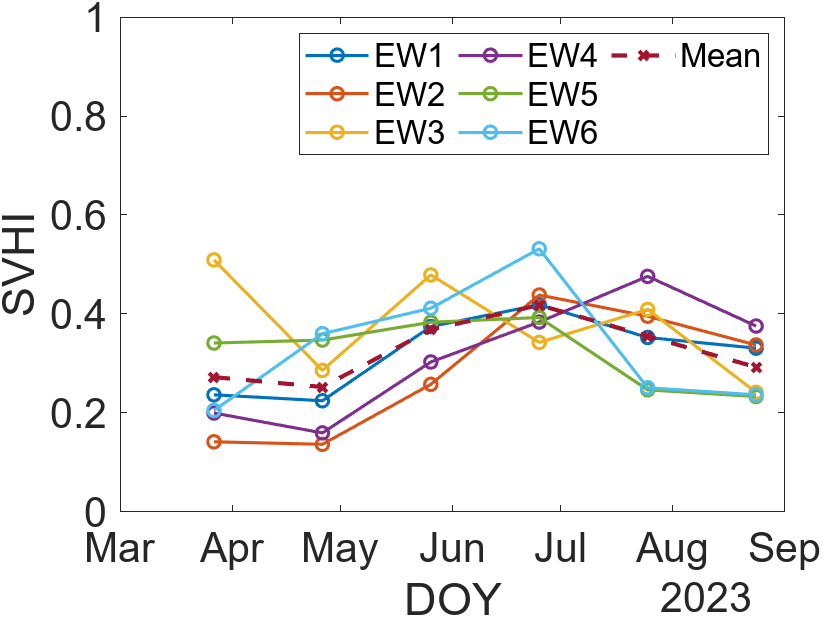}\label{PlotEW_SVHI}}
\hspace{2em}
\subfloat[LAI EW]{%
	\includegraphics[width=0.60\columnwidth]{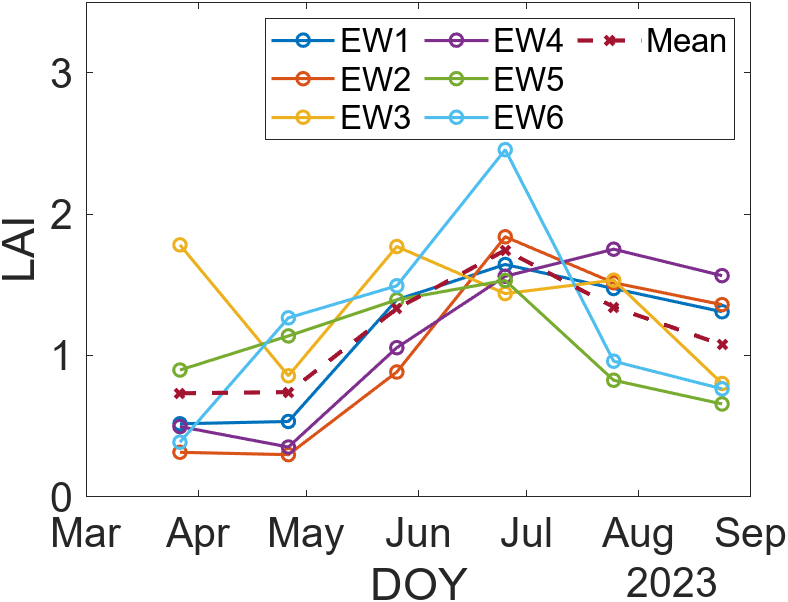}\label{PlotEW_LAI}} 
\caption{Behaviour, as a function of day of the year, of the vegetation indices for the East-West oriented vineyards.}   
\label{fig:TemporalTrendEW}
\end{figure*}

GDD are calculated as follows \cite{mcmaster1997growing}:

\begin{equation}
    \text{GDD}=\frac{T_{\max}+T_{\min}}{2} - T_{\text{base}}
\end{equation}
where $T_{\max}$ is the daily maximum air temperature, $T_{\min}$ is the daily minimum air temperature, and $T_{\text{base}}$ is the baseline temperature.

Castelan-Estrada et al.~\cite{castelan2002allometric} observed a monotonic increase in vineyard biomass throughout the growing season. This trend aligns with the well-established relationship between vine development and temperature, as vine growth rates are strongly influenced by thermal conditions~\cite{jackson2020wine}.

Since precise temperature measurements are not available for the vineyards under investigation, these data are retrieved from a nearby weather station and used to calculate the cumulative degree days. The temperature measurements were acquired from the \emph{Agenzia Regionale per la Protezione Ambientale} (ARPA) at the Broni weather station (coordinates: \ang{45;2;40.128}N, \ang{9;13;25.9}W), located approximately 6 km from the center of the study area.

Biomass (BB) is related to CDD according to a relationship in the form $\text{BB}=k_{biom} \sqrt{\text{CDD}}$, where BB is the biomass density and $k_{biom}$ is a suitable coefficient linked to the features of the plant, and to environmental conditions. Suppose we hypothesize an approximately linear growing trend of temperature in the spring, considering that CDD is an integral of above-threshold daily temperature over the year. In that case, it is reasonable to assume the general trend  $\mathrm{CDD} \approx \mathrm{DoY}^2$ (where DoY stands for day of the year). Therefore, biomass can be approximated as $\text{BB}\approx k_{biom}\, \text{DoY}$. 

In other words, assuming that temperature is only considered during spring and summer and increases linearly with season, biomass can be roughly approximated as a linear function of DoY multiplied by a coefficient or as a parabolic function of CDD multiplied by a coefficient. In any case, DpRVI should behave consistently from March to August to obtain a meaningful result.

\section{Findings and Interpretation}
This section analyzes the SAR- and optical-based vegetation indices for vineyard characterization.

\subsection{Comparison of DpRVI and Optical indices}

\begin{figure*}[h!]
\centering
\subfloat[DpRVI NS, Asc.]{%
	\includegraphics[width=0.60\columnwidth]{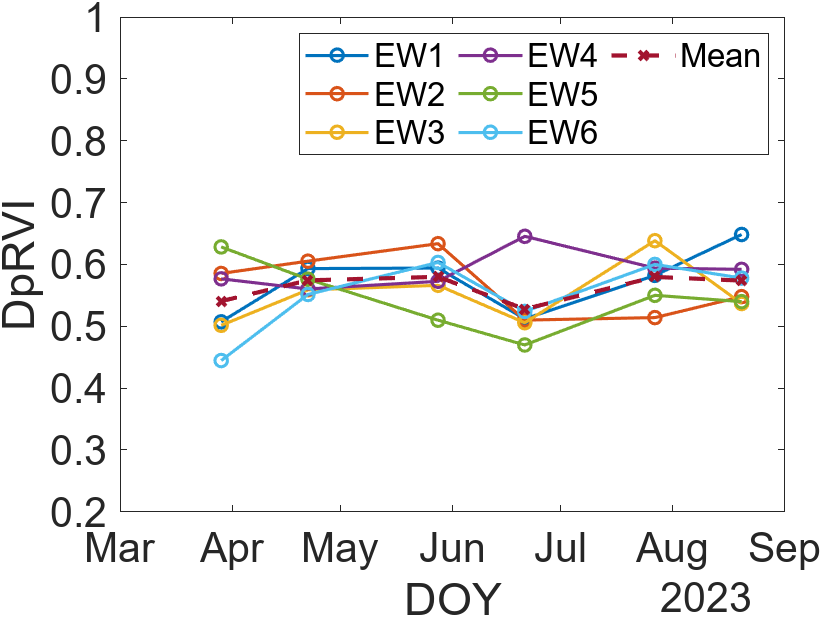}\label{PlotNS_DpRVI_asc}}
\hspace{2em}
\subfloat[DpRVI NS, Des.]{%
	\includegraphics[width=0.60\columnwidth]{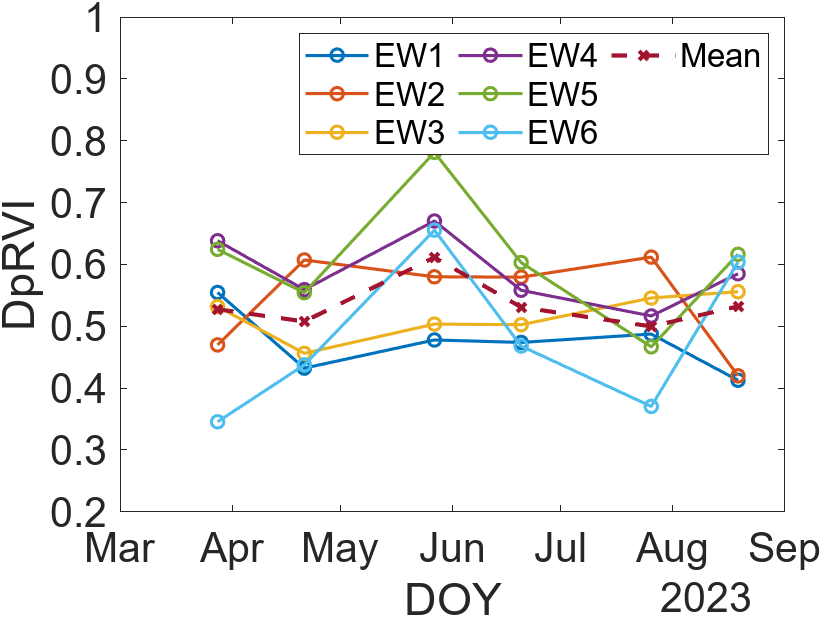}\label{PlotNS_DpRVI_des}}
\hspace{2em}
\subfloat[NDVI NS]{%
	\includegraphics[width=0.60\columnwidth]{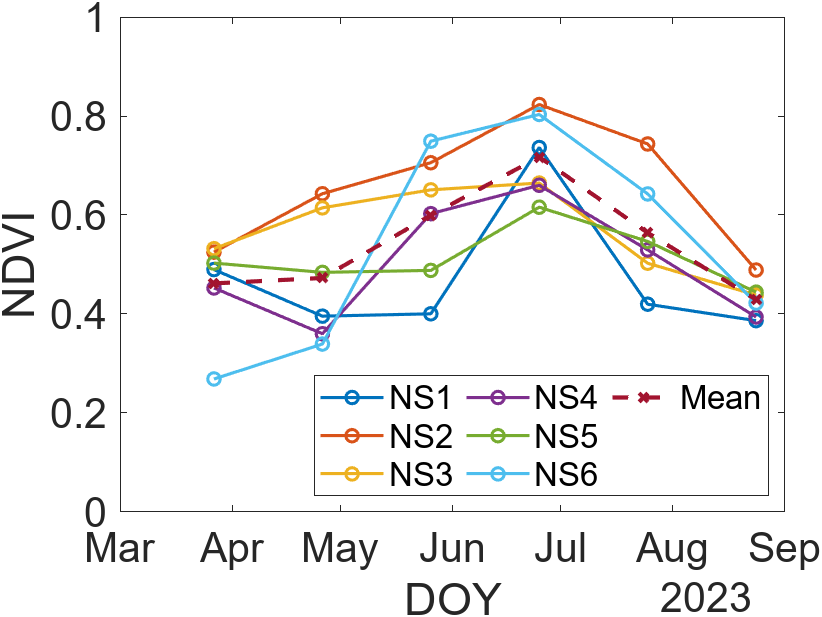}\label{PlotNS_NDVI}}
\\
\subfloat[SVHI NS]{%
	\includegraphics[width=0.60\columnwidth]{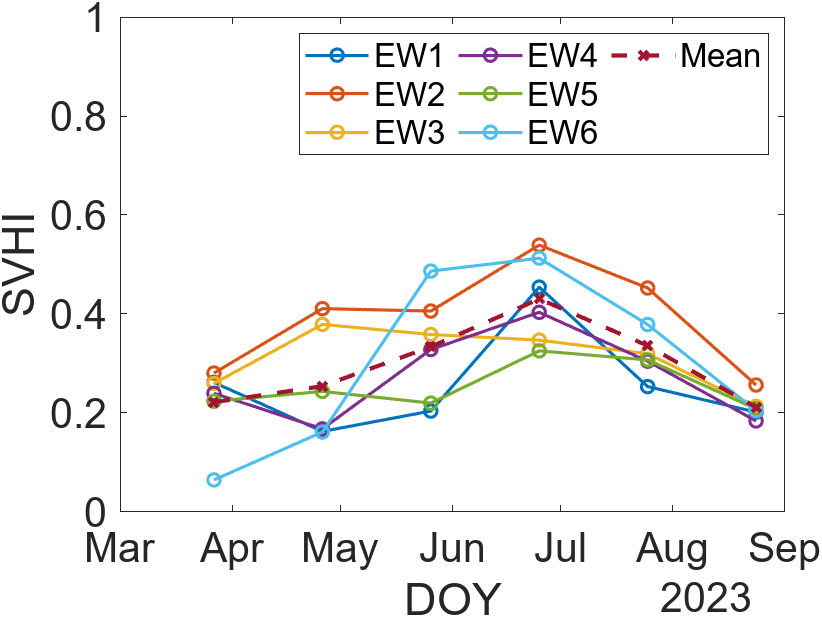}\label{PlotNS_SVHI}}
\hspace{2em}
\subfloat[LAI NS]{%
	\includegraphics[width=0.60\columnwidth]{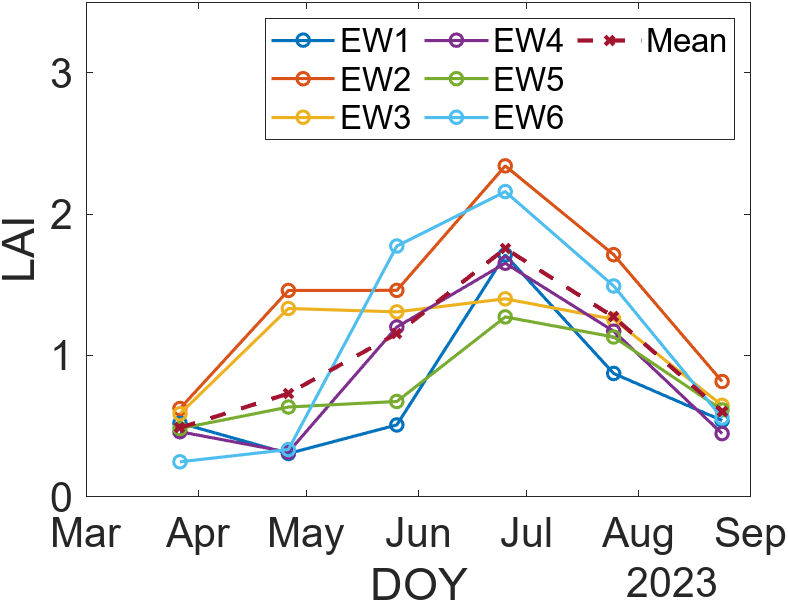}\label{PlotNS_LAI}} 
\caption{Behaviour, as a function of day of the year, of the vegetation indices for the North-South oriented vineyards.}   
\label{fig:TemporalTrendNS}
\end{figure*}

The evolution of DpRVI (for both the ascending and descending orbits), NDVI, SVHI, and LAI is shown as a function of DOY in Figures~\ref{fig:TemporalTrendEW} and \ref{fig:TemporalTrendNS}. The mean values evaluated within each parcel are plotted for both the East-West (Figure~\ref{fig:TemporalTrendEW}) and North-South (Figure~\ref{fig:TemporalTrendNS}) orientation. 

This study analyzes the temporal evolution of the selected indices, with a particular emphasis on evaluating their mutual similarity. The investigation initially focuses on East-West oriented parcels. 

Concerning optical vegetation indices, it should be noted that their evolution is quite similar for all the time intervals considered. At first, their values are low, then they increase until reaching a peak around June, after which they begin to decrease. In particular, NDVI and SVHI show very similar behaviors, with NDVI having higher values than SVHI. Overall, they are highly correlated ($r=0.98$). This can be explained by the fact that both indices use bands B4 and B8. Although SVHI also uses B5, B11, and B12, these additional bands are probably less sensitive to the chemical composition of the vineyard compared to the first two.

LAI exhibits a greater range of values, from $0.25$ in March to $2.46$ in June. However, all optical indices show similar trends; this is not surprising considering they are all derived from the same optical data set.

On the SAR side, the DpRVI remained relatively stable through the observed period, with values ranging from $0.35$ to $0.78$. Small fluctuations were observed, but none were statistically significant or indicative of common major trends. The general trend of the mean DpRVI is quite similar, with increasing values at the beginning of the growing season. However, the descending data show a greater variability, especially in the mid-to-late period. In both cases, the observed trends do not yield statistically robust and reliable conclusions.

Interestingly, the EW3 vineyard showed high values in the optical measurements in late March, while its SAR measurements were consistent with those of the other vineyards. This strange behavior can be explained by several factors, such as microclimate conditions~\cite{jones2000climate}, pruning activities~\cite{senthilkumar2015effect}, or grape variety~\cite{garcia2009performance}. However, the lack of ground truth data makes it challenging to provide a well-supported explanation.

\begin{figure}[h!]
\centering
\subfloat[DpRVI (ascending orbit) vs. LAI for EW2]{%
	\includegraphics[width=0.8\columnwidth]{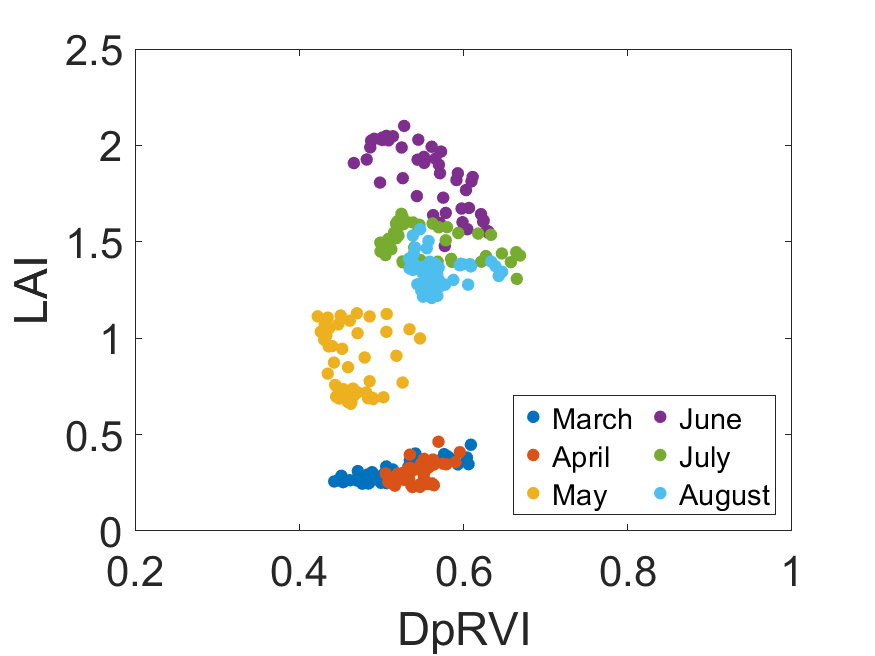}\label{Scatterplot_EW2_Asc}} \\
\subfloat[DpRVI (descending orbit) vs. LAI for EW2]{%
	\includegraphics[width=0.8\columnwidth]{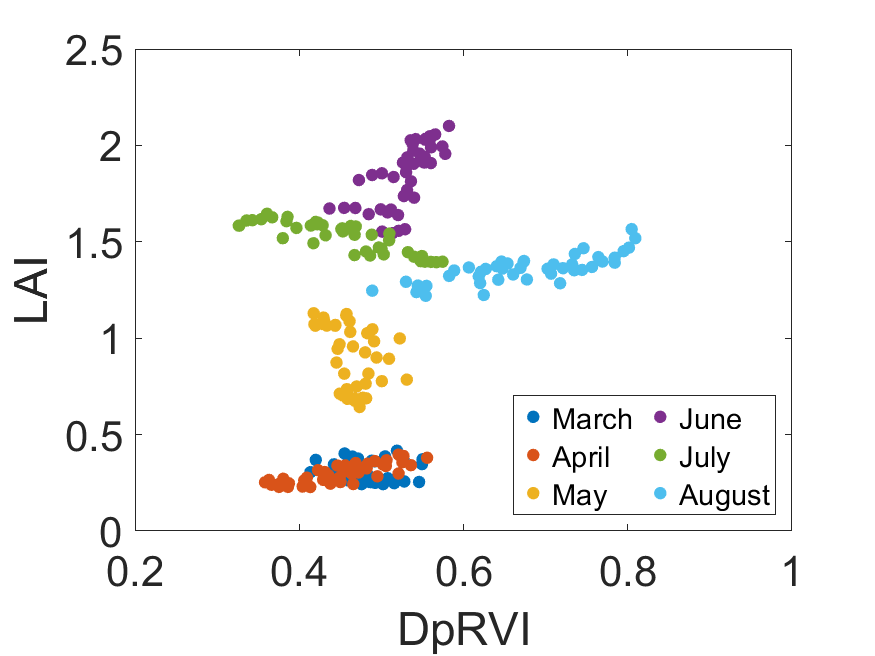}\label{Scatterplot_EW2_Des}} 
\caption{Combined scatter plots between DpRVI and LAI for one sample vineyard EW2.}   
\label{fig:ScatterPlots}
\end{figure}

Regarding the trends for North-South-oriented parcels, NDVI and SVHI show similar evolutions. They exhibit parabolic trends identical to the East-West case but with slightly more variability. These different behaviors may be explained by the fact that the orbit pass, in this case, is roughly parallel to the vineyard row orientation. This may impact the sensitivity and accuracy of the backscatter signal measurement, as it may not interact as strongly with the vine structure.

LAI is the only descriptor that seems to be consistent with its East-West counterpart: in both cases, the peak occurs around June, and the index varies over a wide range of values. LAI describes the vineyard growth pattern, although some variability is observed across the parcels.

Finally, DpRVI, as for the East-West case, does not link to any specific parameter. The index shows more variability and less clear seasonal trends than the optical indices, suggesting that DpRVI is capturing different aspects of the vegetation phenology.

It should be emphasized that the selection of the orbit pass may affect results significantly, as data are acquired at different times of the day depending on the pass, when the vegetative activity may be different.

Scatter plots showing the relationship between DpRVI and LAI over one sample vineyard, namely EW2, are depicted in Figure~\ref{fig:ScatterPlots}. This vineyard was selected as it was representative of all the studied vineyards, which show similar behavior. The data points are colored according to the month they refer to. The analysis was conducted between DpRVI and LAI rather than between DpRVI and other optical indices, as LAI appeared to produce the most consistent results. Because the indices were retrieved from different satellites with varying spatial resolutions, LAI values were selected based on their spatial proximity to DpRVI values. 

Different clusters of data points are clearly visible, suggesting a temporal progression in both DpRVI and LAI. The March and April data points show low LAI and moderate-low DpRVI, which can be attributed to the expected low vegetative activity at this time of year. Both indices tend to increase throughout the growing season, peaking in June and subsequently decreasing. 

The low values in July and August can be explained by leaf senescence and fruit development. Moreover, DpRVI is generally more variable than LAI, and this variability further increases as the year progresses, especially for the descending orbit case. Overall, DpRVI and LAI appear sensitive to different vineyard structure characteristics and biophysical properties.

\subsection{DpRVI for Vineyard Biomass Estimation}

\begin{figure*}[h]
\centering
\subfloat[DpRVI EW4, Asc.]{%
	\includegraphics[width=0.5\columnwidth]{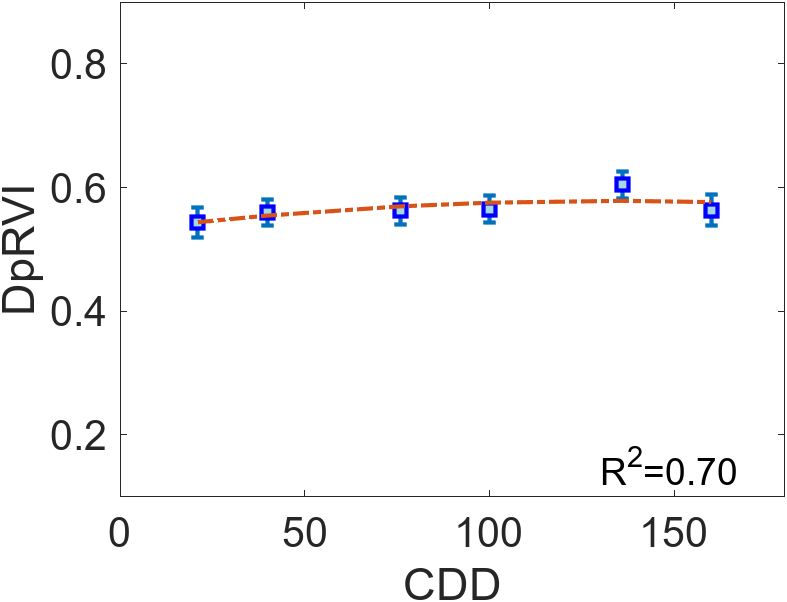}\label{BiomassTrend_EW4_Asc_paper}}
\hfill
\subfloat[DpRVI EW4, Des.]{%
	\includegraphics[width=0.5\columnwidth]{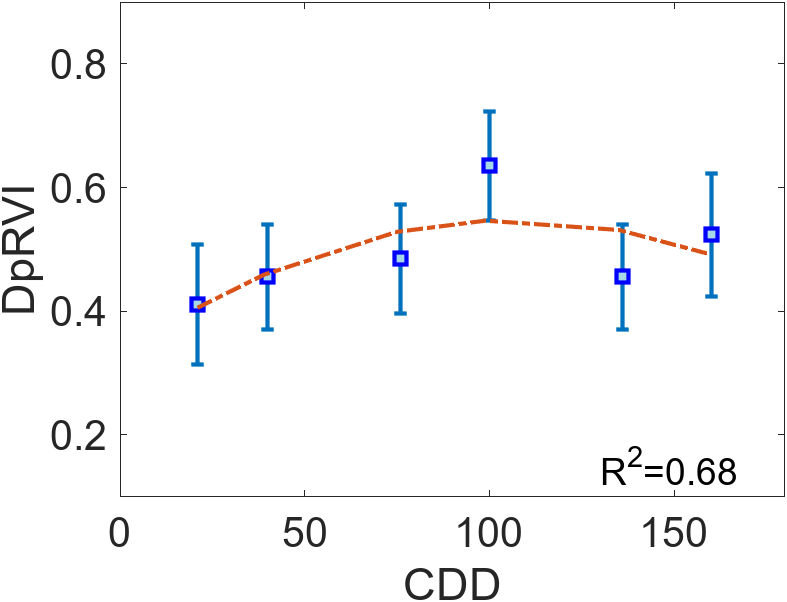}\label{BiomassTrend_EW4_Des_paper}} 
\hfill
\subfloat[DpRVI NS6, Asc.]{%
	\includegraphics[width=0.5\columnwidth]{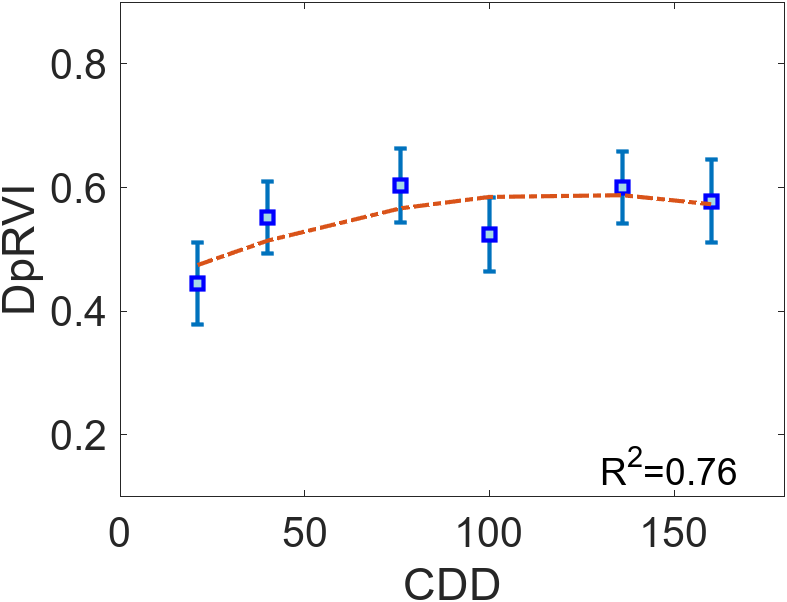}\label{BiomassTrend_NS6_Asc_paper}}
\hfill
\subfloat[DpRVI NS6, Des.]{%
	\includegraphics[width=0.5\columnwidth]{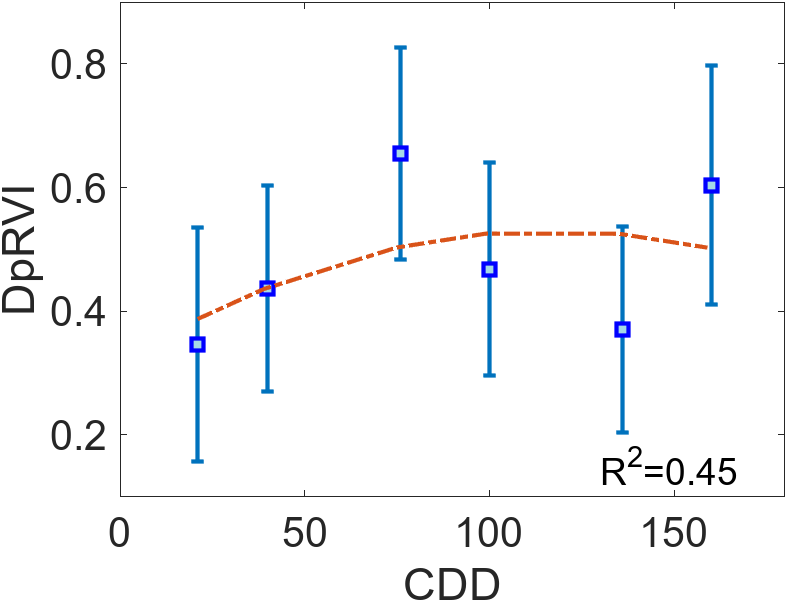}\label{BiomassTrend_NS6_Des_paper}}
\hfill 
\caption{DpRVI evolution as a function of cumulative degree days for two sample vineyards: EW4, NS6.}   
\label{fig:BiomassTrend}
\end{figure*}

This section addresses DpRVI's capacity to describe vineyard biomass. Figure \ref{fig:BiomassTrend} shows the evolution of DpRVI as a function of cumulative degree days, for two sample vineyards (namely, EW4 and NS6) in ascending and descending satellite orbit passes. A summary of the main DpRVI characteristics for all the vineyards studied is presented in Tables \ref{tab:DpRVI_asc} and \ref{tab:DpRVI_des}, whereas Table \ref{tab:DatesConversion} shows the evolution of CDD as a function of DoY in the study area. 

\begin{table}[!h]
    \centering
    \caption{Characteristics of DpRVI for ascending orbit pass.}
    \resizebox{0.90\columnwidth}{!}{%
    \begin{tabular}{ccc}
        \toprule Vineyard ID & DpRVI Peak (Asc) & Parabolic Fit (Asc) \\
        \midrule 
        EW1 & $2023-08-20$ & $0.57$ \\
        EW2 & $2023-08-20$ & $0.62$ \\
        EW3 & $2023-07-27$ & $0.90$ \\
        EW4 & $2023-07-27$ & $0.70$ \\
        EW5 & $2023-07-27$ & $0.80$ \\
        EW6 & $2023-07-27$ & $0.57$ \\
        NS1 & $2023-08-20$ & $0.61$ \\
        NS2 & $2023-05-28$ & $0.62$ \\
        NS3 & $2023-07-27$ & $0.45$ \\
        NS4 & $2023-06-21$ & $0.58$ \\
        NS5 & $2023-03-29$ & $0.93$ \\
        NS6 & $2023-05-28$ & $0.76$ \\
        \bottomrule
    \end{tabular}}
    \label{tab:DpRVI_asc}
\end{table}

\begin{table}[!h]
    \centering
    \caption{Characteristics of DpRVI for descending orbit pass.}
    \resizebox{0.90\columnwidth}{!}{%
    \begin{tabular}{ccc}
        \toprule Vineyard ID & DpRVI Peak (Des) & Parabolic Fit (Des) \\
        \midrule 
        EW1 & $2023-06-20$ & $0.97$ \\
        EW2 & $2023-08-19$ & $0.78$ \\
        EW3 & $2023-04-21$ & $0.77$ \\
        EW4 & $2023-06-20$ & $0.68$ \\
        EW5 & $2023-07-26$ & $0.55$ \\
        EW6 & $2023-05-27$ & $0.69$ \\
        NS1 & $2023-03-28$ & $0.54$ \\
        NS2 & $2023-07-26$ & $0.80$ \\
        NS3 & $2023-08-19$ & $0.78$ \\
        NS4 & $2023-05-27$ & $0.46$ \\
        NS5 & $2023-05-27$ & $0.35$ \\
        NS6 & $2023-05-27$ & $0.45$ \\
        \bottomrule
    \end{tabular}}
    \label{tab:DpRVI_des}
\end{table}

The graphs suggest that DpRVI follows a parabolic evolution. However, the relationship is weak, as indicated by the low $R$-squared values, implying that factors other than biomass significantly impact DpRVI. One such factor may be precipitation: the vineyard absorbs more water in rainy weather, causing the radar signal to penetrate less into the vegetation layers and interact more with the plant canopy, decreasing the volume scattering and the DpRVI value.

\begin{figure}[h]
    \centering
    \includegraphics[width=0.8\linewidth]{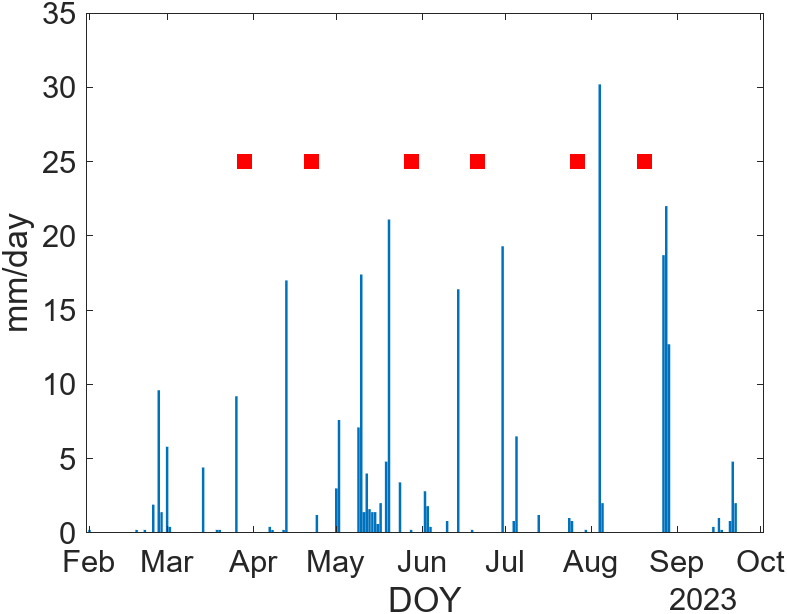}
    \caption{Daily rainfall (mm/day) from 1 Feb to 1 Oct 2023 in the study area. Sentinel-1 ascending acquisition dates are indicated with red markers.}
    \label{fig:rain}
\end{figure}

The precipitation bar graph for the Broni weather station is depicted in Figure \ref{fig:rain}. However, at least visually, DpRVI does not appear to correlate with precipitation peaks. The almost even distribution during heavy rainfall episodes does not help identify its impact on the radar vegetation index.

As seen from Table \ref{tab:DpRVI_asc}, the ascending acquisitions show the best parabolic fits, with an average Pearson correlation coefficient of $0.72$ for EW and $0.66$ for NS. The location of the DpRVI peaks also differs between the two types of passes, with descending peaks generally occurring earlier than ascending ones. 

\begin{table}[!h]
    \centering
    \caption{Cumulative degree days recorded on each DoY with relevant satellite acquisitions.}
    \resizebox{0.45\columnwidth}{!}{%
    \begin{tabular}{c c }
        \toprule DoY & CDD \\
        \midrule 
        $2023-03-28$ & $21$ \\
        $2023-03-29$ & $21$ \\
        $2023-04-21$ & $39$ \\
        $2023-04-22$ & $40$ \\
        $2023-05-27$ & $75$ \\
        $2023-05-28$ & $76$ \\
        $2023-06-20$ & $99$ \\
        $2023-06-21$ & $100$ \\
        $2023-07-26$ & $135$ \\
        $2023-07-27$ & $136$ \\
        $2023-08-19$ & $159$ \\
        $2023-08-20$ & $160$ \\
        \bottomrule
    \end{tabular}}
    \label{tab:DatesConversion}
\end{table}

\section{Conclusions and future work}

In this study, for the first time, a radar vegetation index derived from C-band dual-pol Sentinel-1 SLC SAR data is compared with optical indices for monitoring a vineyard's seasonal development during the agricultural year. The DpRVI index that serves as a proxy for vegetation characteristics is derived using the eigenvalue spectrum of the $2\times 2$ covariance matrix. Three experimental analyses have been conducted to assess different properties of the DpRVI for vineyard characterization. 

Experimental results show that the DpRVI is not correlated with optical indices (NDVI and SVHI), suggesting that the indices capture different aspects of vineyard phenology. Additionally, DpRVI, evaluated based on cumulative degree days, follows a similar parabolic evolution through the vineyard growing season, suggesting its potential as a starting point for vineyard biomass estimation.

This research regards the joint application of SAR and optical data to seasonal analysis of vineyards, a research field largely unexplored in the scientific literature. For this reason, several improvements can be made, such as:

\begin{itemize}
    \item Combining C-band Sentinel-1 data with X-band data (e.g., COSMO-SkyMed) to enhance the radar sensitivity to vineyard canopy, as X-band waves penetrate less deeply into vegetation.
    \item Using fully polarimetric SAR data to discriminate scattering mechanisms more effectively.
    \item Applying phenological modeling (such as those used in~\cite{mascolo2016complete} and~\cite{mascolo2015retrieval}) to improve biomass monitoring and estimation of phenological phases.
    \item Introducing ground-truth measurements (e.g., LIDAR, multispectral sensors) to validate the results of satellite observations.
\end{itemize}


\bibliographystyle{IEEEtran}
\bibliography{Bibliography.bib}

\EOD

\end{document}